\documentclass[10pt]{article}
\usepackage[preprint]{tmlr}
\usepackage{amsmath}
\usepackage{amsfonts}
\usepackage{booktabs}
\usepackage{graphicx}
\usepackage{microtype}
\usepackage{xcolor}
\usepackage{url}
\usepackage{hyperref}
\usepackage{float}
\usepackage{multirow}
\usepackage{wrapfig}
\usepackage{caption}
\usepackage{subcaption}
\newif\ifcamera
\cameratrue % arXiv

\title{VQ-Atom: Semantic Discretization of Local Atomic Environments for Molecular Representation Learning}
\ifcamera
\author{
Takayuki Kimura\\
Atoms as Language, LLC\\
\texttt{info@atomsaslanguage.com}
}
\else
\author{}
\fi

\begin{document}

\maketitle

\begin{abstract}
Large language models succeed by combining large-scale pretraining with meaningful discrete tokens. In molecular machine learning, SMILES is widely used as a token representation, but it is primarily a linearization format for molecular graphs rather than a semantic decomposition of chemistry. We propose VQ-Atom, a semantic tokenization framework that assigns discrete atom-level tokens based on local chemical environments via vector quantization. Unlike SMILES tokens, VQ-Atom tokens encode graph-local chemical context and are aligned with molecular structure. On protein-cold drug--target interaction prediction using the KIBA dataset, VQ-Atom substantially improves global ranking performance, achieving AUROC of 0.79 while substantially outperforming both SMILES-based and continuous molecular representations under an identical downstream architecture. Furthermore, VQ-Atom enables approximately 3 times faster downstream training than continuous atom-level representations by replacing per-atom continuous features with reusable discrete tokens. These results suggest that molecular tokenization is not merely a preprocessing step, but a central design choice. In particular, well-structured tokens can encode substantial chemical semantics, reducing the burden on downstream learning. VQ-Atom can be interpreted as defining a molecular language, where tokens correspond to chemically meaningful atomic environments, suggesting that token design may constitute an additional axis of machine learning research alongside architecture, objectives, and optimization.

\end{abstract}

\section{Introduction}

Transformer-based models have achieved remarkable success across domains ranging from natural language processing to biology and chemistry~\cite{devlin2019bert,brown2020language}. While much attention has been devoted to model architecture and large-scale pretraining, an equally important factor is the design of the underlying representation. In natural language, Transformer models operate on discrete tokens such as words or subwords that provide stable symbolic units over which statistical regularities can be accumulated~\cite{sennrich2016bpe}. The effectiveness of pretraining therefore depends not only on model capacity, but also on the quality of the tokenization itself.

\begin{figure}[t]
\centering
\includegraphics[width=0.85\linewidth]{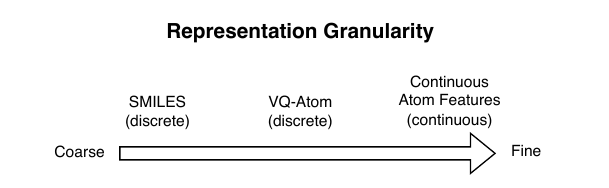}
\caption{
Conceptual comparison of molecular representations in terms of representation granularity. SMILES provides discrete but relatively coarse symbolic units, while continuous atom-level features provide fine-grained information but remain continuous. VQ-Atom occupies an intermediate position by preserving atom-level granularity while converting local chemical environments into reusable discrete tokens.
}
\label{fig:granularity}
\end{figure}

Molecular machine learning has increasingly adopted Transformer-based approaches, yet there is no consensus regarding the appropriate molecular representation. Existing approaches largely fall into two categories (Figure~\ref{fig:granularity}).

The first uses SMILES~\cite{weininger1988smiles}, a string representation that naturally provides discrete tokens compatible with Transformer architectures. Transformer pretraining on SMILES has been extensively studied through masked language modeling and related objectives on large molecular corpora~\cite{chithrananda2020chemberta,fabian2020molbert,ross2022molformer}. However, these approaches inherit the tokenization induced by SMILES, whose tokens are primarily syntactic symbols derived from graph traversal and often fail to reflect chemically meaningful local environments. As a result, they do not address whether the tokenization itself is optimal for molecular learning.

The second category uses graph neural networks (GNNs)~\cite{gilmer2017neural}, which generate continuous atom-level embeddings that preserve rich chemical information. While these representations provide substantially finer granularity than SMILES, they remain continuous rather than symbolic. Consequently, chemically similar local environments are not mapped to reusable discrete units, limiting the accumulation of statistical evidence across molecules that underlies token-based Transformer pretraining~\cite{sennrich2016bpe,devlin2019bert,brown2020language}.

% This observation suggests a fundamental trade-off between two desirable properties of molecular representations: granularity and discreteness. SMILES provides reusable discrete symbols but operates at a relatively coarse granularity, whereas continuous atom embeddings preserve fine-grained chemical information but remain non-discrete. VQ-Atom is motivated by the hypothesis that both properties are necessary for effective Transformer-based molecular learning.

\begin{figure*}[t]
\centering
\includegraphics[width=\textwidth]{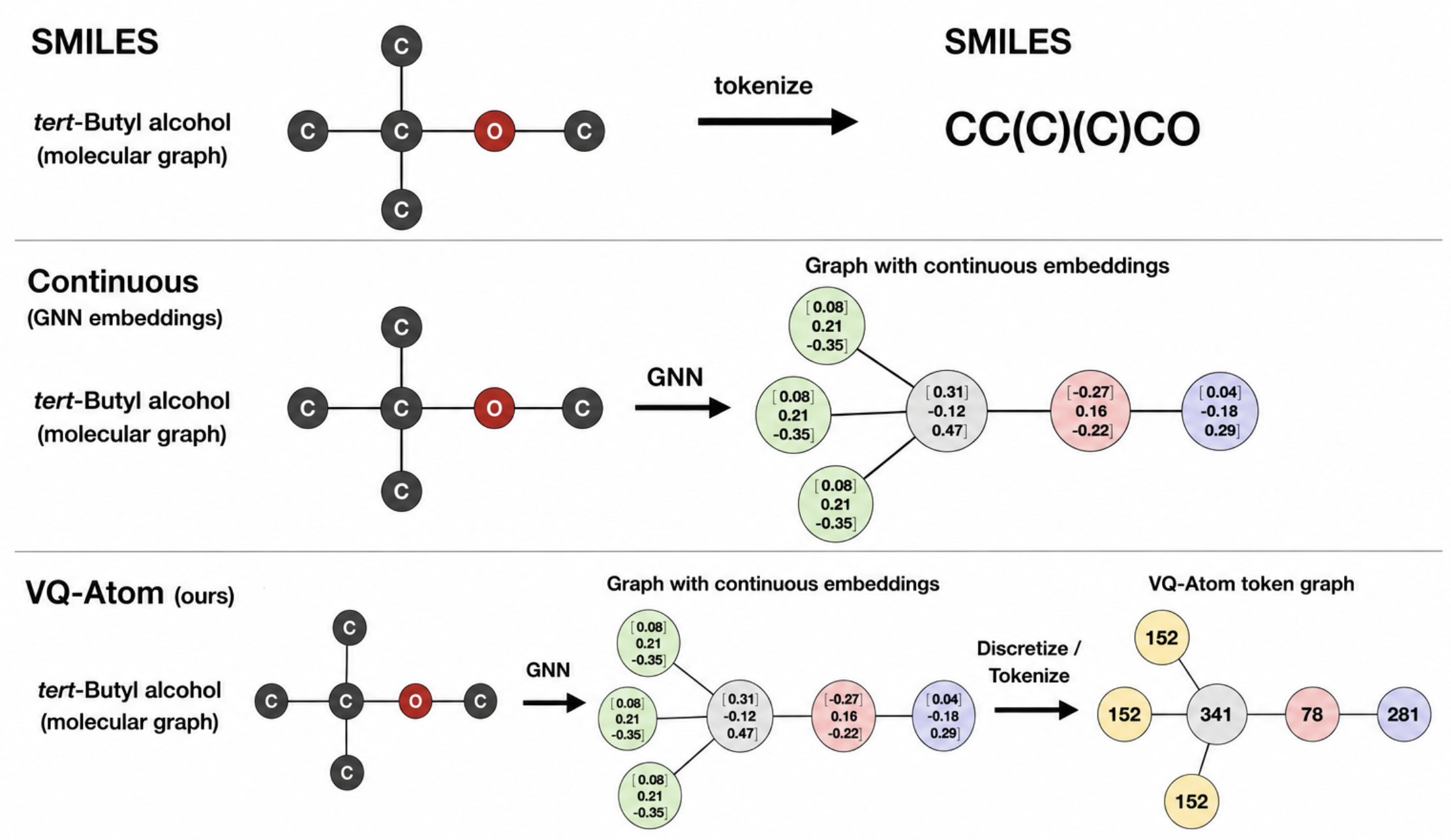}
\caption{
Comparison of molecular representation pipelines.
SMILES directly converts a molecular graph into a sequence of discrete symbols through graph linearization and tokenization.
Continuous representations apply a graph neural network (GNN) to generate atom-level continuous embeddings while preserving graph structure.
VQ-Atom combines both approaches: a GNN first produces continuous atom representations, which are subsequently discretized into reusable semantic tokens through vector quantization.
As a result, equivalent local chemical environments can be assigned identical token IDs across molecules while retaining the fine-grained locality of graph-based representations.
}
\label{fig:three_represent}
\end{figure*}

Figure~\ref{fig:granularity} illustrates a fundamental trade-off between two desirable properties of molecular representations: granularity and discreteness.
SMILES provides reusable discrete symbols but operates at a relatively coarse granularity, whereas continuous atom embeddings preserve fine-grained chemical information but remain non-discrete. VQ-Atom is motivated by the hypothesis that both properties are necessary for effective Transformer-based molecular learning. Figure~\ref{fig:three_represent} compares the three representation paradigms studied in this work.

Recent work has explored vector quantization (VQ) as a mechanism for learning discrete latent representations in images and generative modeling~\cite{oord2017vqvae,razavi2019vqvae2}. Similar ideas have recently been applied to molecular generation, where VQ-based models learn discrete latent codes for autoregressive or diffusion-based molecule generation~\cite{zheng2025gvt,noravesh2026vqsad}. However, these approaches primarily use discrete codes as latent variables for reconstruction or generation. In contrast, VQ-Atom uses vector quantization to construct a reusable atom-level vocabulary, where token identities correspond to recurring local chemical environments and serve directly as inputs for Transformer-based representation learning. Recent work has explored vector quantization (VQ) as a mechanism for learning discrete latent representations in images and generative modeling~\cite{oord2017vqvae,razavi2019vqvae2}. Similar ideas have recently been applied to molecular generation, including graph-based latent tokenization approaches such as GVT~\cite{zheng2025gvt} and VQSAD~\cite{noravesh2026vqsad}, where discrete codes are learned for autoregressive or diffusion-based molecule generation. However, these approaches primarily use discrete codes as latent variables for reconstruction or generation. Several recent studies have explored atom-level discretization and vector-quantized molecular representations for downstream learning and generation.

In this work, we propose \textbf{VQ-Atom}, a semantic tokenization framework designed to combine both properties. Starting from atom-centered local chemical environments, we learn a vector-quantized vocabulary in which each token corresponds to a recurring chemical environment. The resulting representation preserves the fine-grained information available in graph-based molecular encodings while converting it into reusable discrete symbols. In this sense, VQ-Atom can be viewed as defining a molecular vocabulary whose units correspond to chemically meaningful atomic environments rather than syntactic fragments.

To evaluate this hypothesis, we compare three representation paradigms under an identical downstream architecture: SMILES, continuous atom-level embeddings, and VQ-Atom. We perform experiments on protein-cold drug--target interaction prediction using the KIBA benchmark, a challenging setting requiring generalization to unseen protein families.

Our results reveal a consistent pattern. SMILES performs poorly regardless of pretraining, suggesting that discreteness alone is insufficient. Continuous atom representations substantially outperform SMILES, demonstrating the importance of fine-grained chemical information, but benefit only marginally from Transformer pretraining. These findings suggest that effective molecular representations require both granularity and discreteness, and that tokenization should be viewed as a central modeling decision rather than a preprocessing detail.

The contributions of this work are threefold:

\begin{itemize}
\item We identify a fundamental granularity--discreteness trade-off in molecular representation learning for Transformer-based models.

\item We introduce VQ-Atom, a semantic discretization framework that converts atom-level chemical environments into reusable discrete tokens.

\item We demonstrate that VQ-Atom enables substantially more effective Transformer training than either SMILES or continuous atom representations, achieving the best overall performance on protein-cold DTI prediction.
\end{itemize}

\begin{figure}[t]
\centering
\includegraphics[width=0.95\linewidth]{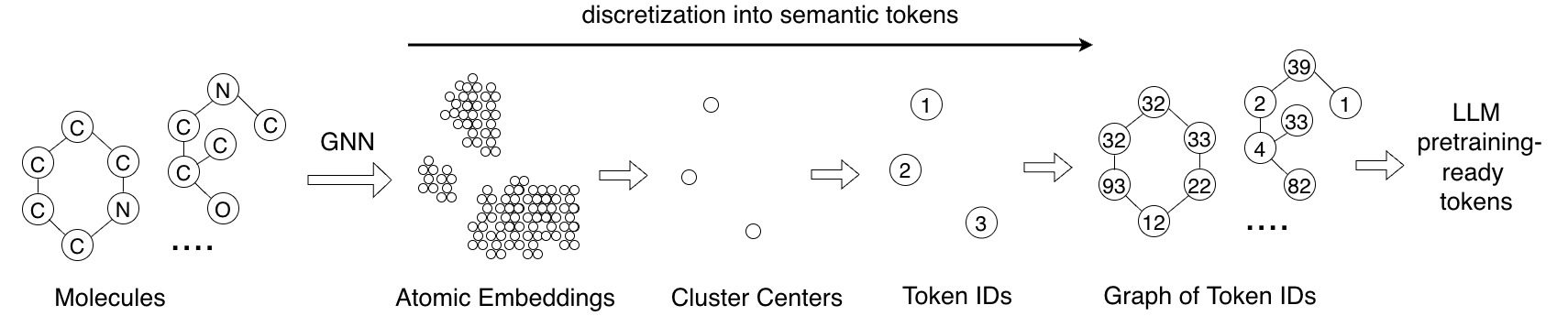}
\caption{
VQ-Atom framework. 
Molecular graphs are first encoded into atom-level embeddings using a graph neural network (GNN), capturing local chemical environments. 
These continuous embeddings are then discretized via vector quantization into a finite set of codebook vectors (cluster centers). 
Each atom is assigned the index of its nearest codebook vector, forming discrete token IDs. 
The resulting graph of token IDs serves as input to Transformer-based pretraining.
}
\label{fig:vqatom}
\end{figure}

\begin{figure}[t]
\centering
\includegraphics[width=0.98\linewidth]{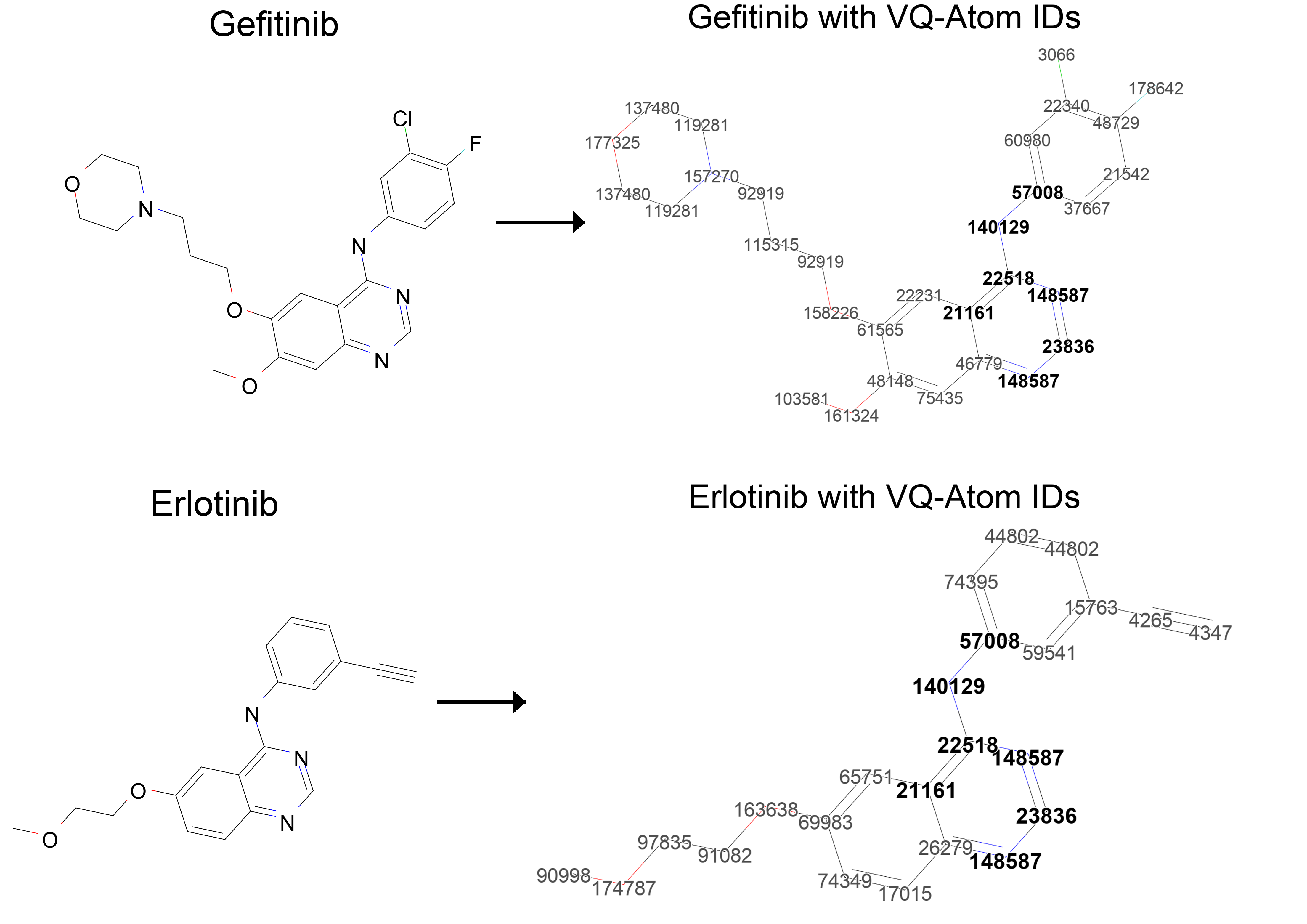}
\caption{
Examples of VQ-Atom tokenization. 
\textbf{Left:} original molecular structures (Gefitinib and Erlotinib). 
\textbf{Right:} corresponding VQ-Atom representations, where each atom is assigned a discrete token ID based on its local chemical environment. 
Identical local environments are mapped to the same token ID across molecules. Bolded IDs highlight recurring token patterns shared across chemically similar substructures.
}
\label{fig:sample_molecule}
\end{figure}

\section{VQ-Atom: Overview}

Figure~\ref{fig:vqatom} illustrates the overall pipeline, from continuous atom-level representations to discrete semantic tokens. VQ-Atom constructs a discrete molecular vocabulary by encoding atom-centered local chemical environments and discretizing them via vector quantization.

Each atom is represented by a feature vector capturing its local chemical context, including atomic identity, degree, formal charge, hybridization, aromaticity, ring membership, hydrogen count, and functional group indicators~\cite{rogers2010extended}. We further include higher-order descriptors such as ring size, fused ring identity, aromatic neighbor counts, and bond-environment features. These features are encoded by a graph neural network to produce atom-level representations.

Vector quantization~\cite{oord2017vqvae} maps these continuous embeddings into a finite set of discrete codes. Unlike SMILES~\cite{weininger1988smiles}, where token identity depends on a particular graph traversal, VQ-Atom tokens are assigned directly to atom-centered local environments and are invariant to linearization. The resulting token sequences are compatible with Transformer-based masked language modeling~\cite{devlin2019bert,brown2020language}, while preserving chemically meaningful local structure. In this formulation, sequence order becomes an implementation detail, whereas chemical context is embedded directly into the tokens.

Figure~\ref{fig:sample_molecule} shows examples of this transformation, where similar local environments are mapped to shared token IDs across molecules, while distinct environments receive different tokens. These observations indicate that VQ-Atom captures chemically meaningful local structure while enabling reuse of common patterns across molecules. Together, Figures~\ref{fig:vqatom} and \ref{fig:sample_molecule} illustrate how recurring chemical environments are mapped to identical token IDs across distinct molecules. This enables statistical accumulation over shared chemical motifs, analogous to how words are reused across different sentences.

Our goal is not to introduce a novel graph neural network architecture, but to study the effect of semantic discretization. Therefore, we keep the encoder simple and focus on the properties of the resulting tokenization.
\section{Method}

\subsection{VQ-Atom Discretization}
Vector quantization has previously been explored in generative modeling, where discrete latent codes are learned for images, language, and molecular structures~\cite{oord2017vqvae,razavi2019vqvae2,ramesh2021dalle}. Prior molecular VQ approaches primarily use discrete codes as latent variables for reconstruction or generation~\cite{zheng2025gvt}. In contrast, VQ-Atom employs vector quantization as a semantic tokenization mechanism, with the resulting codes serving directly as reusable atom-level vocabulary units for Transformer-based representation learning.

Unlike reconstruction-oriented VQ methods, VQ-Atom does not use discrete codes as latent variables for generation. Instead, the learned codebook defines a molecular vocabulary whose entries are used directly as atom-level tokens.

Each molecule is represented as an atom-level graph and encoded using a three-layer GINE network with jumping-knowledge aggregation over 0--3 hop neighborhoods. Atom features include elemental identity, degree, formal charge, hybridization, aromaticity, ring membership, hydrogen count, functional-group indicators, donor/acceptor flags, ring size, aromatic neighbor count, fused-ring identity, and local bond-environment descriptors.

The encoder maps each atom to a 16-dimensional latent representation followed by L2 normalization. Atoms are grouped according to intrinsic chemical properties (atomic number, charge state, hybridization, aromaticity, ring membership, and hydrogen count), and a separate vector-quantization codebook is maintained for each group. Codebooks are initialized using K-means and subsequently updated using exponential moving averages (EMA).

Our objective focuses on learning stable and semantically meaningful atom-environment clusters. The quantization module is trained using a commitment loss together with an entropy-based regularization term:

\begin{equation}
\mathcal{L}_{\mathrm{VQ}}
=
\mathcal{L}_{\mathrm{commit}}
-
\lambda_{\mathrm{ent}}
\mathcal{L}_{\mathrm{ent}} .
\end{equation}

The commitment loss encourages atom embeddings to align with their assigned codebook vectors, while the entropy term encourages balanced utilization of the codebook and mitigates collapse to a small subset of tokens.

\begin{equation}
\mathcal{L}_{\mathrm{ent}}
=
-\sum_{k=1}^{K}
p_k \log p_k ,
\end{equation}
where $p_k$ denotes the empirical assignment frequency of codebook entry $k$.
After training, the codebooks are frozen and each atom is assigned to its nearest codebook entry using cosine similarity. The resulting codebook index becomes the VQ-Atom token ID. In this way, chemically similar local environments are mapped to shared discrete tokens while preserving fine-grained structural information.

The codebooks are trained on approximately 208k unlabeled molecular structures derived primarily from ChEMBL~\cite{gaulton2012chembl}. The discretization stage uses molecular structures only and does not use protein--ligand interaction labels from the downstream KIBA benchmark~\cite{he2017simboost}.labels.

Our goal is not to optimize molecular reconstruction accuracy, but rather to learn a reusable molecular vocabulary. 

Consequently, the training objective encourages stable and diverse token assignments, yielding semantically meaningful atom-level tokens suitable for Transformer-based learning.

\subsection{Ligand Pretraining (MLM)}

We pretrain a Transformer-based ligand encoder on VQ-Atom token sequences using a masked language modeling (MLM) objective. Molecules are represented as variable-length token sequences and padded within each batch. We pretrain on approximately 1.0 million unlabeled molecules (1,007,434 molecules in total), comprising about 28 million atom-level tokens. Tokens are randomly masked with probability 0.15 following the standard 80/10/10 strategy. The model is trained to predict the original token identity using a cross-entropy loss over the discrete VQ-Atom vocabulary.
\begin{equation}
\mathcal{L}_{\text{MLM}} = -\mathbb{E}_{(x,i)} \log p_\theta(x_i \mid x_{-i})
\end{equation}
where $x_i$ denotes the token at position $i$, and $x_{-i}$ denotes the input with the $i$-th token masked. The encoder consists of 6 Transformer layers with hidden dimension 256, 8 attention heads, and feed-forward dimension 1024, with dropout set to 0.1. Training is performed using AdamW with learning rate $2\times10^{-4}$, weight decay 0.01, batch size 64, gradient clipping at 1.0, and a warmup--cosine learning-rate schedule. We train the model for up to 65 epochs and select the best checkpoint based on validation loss. The final pretrained model (epoch 5) is used for downstream experiments. The MLM objective converges stably after correcting the tokenization pipeline. The MLM preprocessing pipeline, split configuration, and training scripts will be released.

\begin{figure}[t]
\centering
\includegraphics[width=\linewidth]{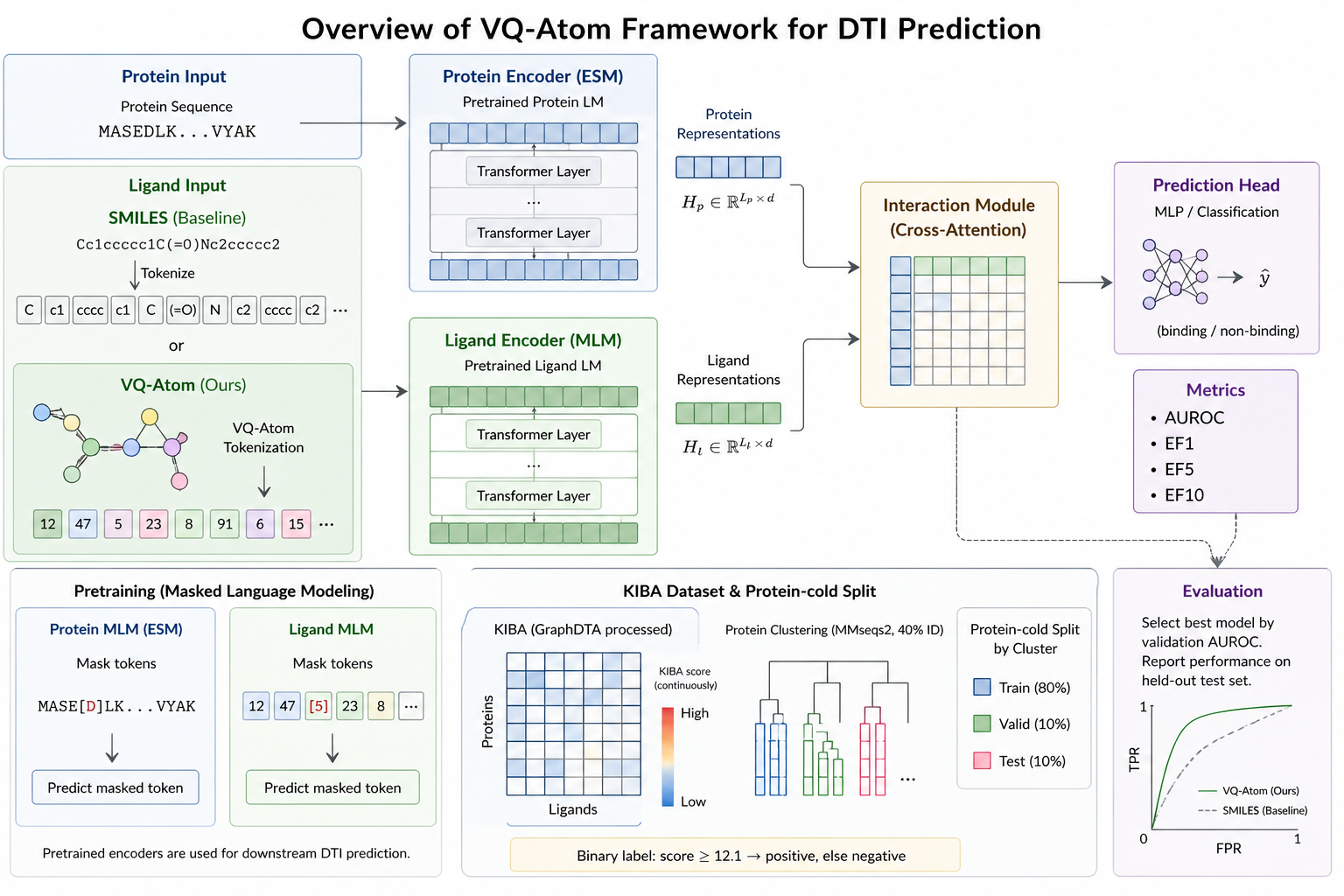}
\caption{
Overview of the DTI prediction framework. Protein sequences are encoded using a pretrained ESM model, while ligands are represented using either SMILES or VQ-Atom tokens. Both ligand representations are processed with the same Transformer-based encoder and fused via a cross-attention interaction module. The only difference between the two settings is the ligand tokenization, enabling a controlled comparison of representation quality.
}
\label{fig:dti_framework}
\end{figure}

\subsection{DTI Model}

Token-based ligand representations (SMILES and VQ-Atom) are encoded using a Transformer encoder with six layers, hidden dimension 256, and eight attention heads. Continuous atom-level representations are encoded using a graph neural network with the same hidden dimensionality. While the ligand encoders differ according to the representation type, the downstream interaction module, prediction head, optimization procedure, and evaluation protocol are shared across all settings.

For downstream drug--target interaction (DTI) prediction, we employ a shared interaction architecture in order to isolate the effect of molecular representation. Proteins are encoded using the pretrained ESM-2 model~\cite{lin2023esm2}, while ligands are represented using one of three alternatives: SMILES tokens, continuous atom-level features, or VQ-Atom tokens. The bottom 28 layers of ESM are frozen and the remaining parameters are fine-tuned during downstream training. The encoder architecture is chosen to match the structure of each representation: sequence-based encoders for tokenized representations and graph-based encoders for continuous atom-level representations.

For token-based ligand representations (SMILES and VQ-Atom), ligands are processed using a Transformer encoder with six layers, hidden dimension 256, and eight attention heads. For continuous representations, atom-level features are encoded using a graph neural network with the same hidden dimensionality. To ensure a fair comparison, all downstream interaction modules and training procedures are kept identical across representations.

We do not use positional encodings for ligand tokens. Although ligand representations are processed as sequences for compatibility with Transformer architectures, neither VQ-Atom tokens nor atom-level molecular representations possess a natural one-dimensional ordering analogous to language. Introducing positional encodings would therefore impose an arbitrary ordering unrelated to molecular structure and could introduce undesirable inductive bias.

Figure~\ref{fig:dti_framework} illustrates the overall architecture. The protein and ligand encoders are fused using a symmetric dual-stream cross-attention mechanism. Given protein representations (P) and ligand representations (L), two directional interaction maps are computed simultaneously: ligand-to-protein attention and protein-to-ligand attention. Query and key vectors are first L2-normalized, resulting in temperature-scaled cosine-similarity attention scores.

Unlike standard Transformer attention, we replace softmax normalization with a sigmoid activation:

\[
A=\sigma\!\left(\frac{QK^\top}{\tau}\right)
\]

A ligand atom can interact with multiple amino-acid residues, and a residue can likewise interact with multiple ligand atoms. Since molecular recognition often involves such many-to-many interactions, we avoid the competitive normalization imposed by softmax and instead evaluate each ligand--protein pair independently using a sigmoid activation. $\tau$ is a temperature parameter controlling the sharpness of the pairwise interaction scores. This formulation allows each ligand--protein pair to be evaluated independently rather than forcing competition among all residues or atoms. Such behavior is more suitable for molecular recognition, where multiple simultaneous contacts may contribute to binding. The final interaction map is obtained by retaining only interactions supported from both directions:

\[
A_{\text{both}}
=
\min\!\left(
A_{L\rightarrow P},
A_{P\rightarrow L}^{\top}
\right)
\]

This overlap operation retains only interactions that are supported from both directions and suppresses one-sided attention artifacts. The resulting symmetric pairwise interaction map is subsequently processed by a lightweight convolutional interaction head to predict binding probability.

Our comparisons focus on molecular representations under a shared downstream architecture. We intentionally avoid comparisons with specialized state-of-the-art DTI architectures because such comparisons would confound the effects of representation learning and architectural design. The goal of this work is not to establish a new state-of-the-art DTI predictor, but rather to evaluate how molecular tokenization influences Transformer-based learning and downstream generalization.

\subsection{DTI Training}

We formulate DTI prediction using both binarized interaction labels and continuous KIBA scores. Continuous KIBA scores are binarized using a threshold of 12.1, and the model outputs a scalar classification logit $z$ as well as a regression prediction $\hat{y}$.

The binary classification term is defined as

\begin{equation}
\mathcal{L}_{\mathrm{BCE}}
=
-\mathbb{E}_{(x,y)}
\left[
y \log \sigma(z)
+
(1-y)\log(1-\sigma(z))
\right],
\end{equation}

where $y\in\{0,1\}$ denotes the binarized interaction label and $\sigma$ denotes the sigmoid function. The regression loss $\mathcal{L}_{\mathrm{reg}}$ is implemented as Smooth L1 loss on the continuous KIBA score, and $\mathcal{L}_{\mathrm{rank}}$ is a pairwise ranking loss computed within protein groups.

The final training objective is

\begin{equation}
\mathcal{L}_{\mathrm{DTI}}
=
\lambda_{\mathrm{cls}}\mathcal{L}_{\mathrm{BCE}}
+
\lambda_{\mathrm{reg}}\mathcal{L}_{\mathrm{reg}}
+
\lambda_{\mathrm{rank}}\mathcal{L}_{\mathrm{rank}}
+
\lambda_{\mathrm{contact}}\mathcal{L}_{\mathrm{contact}} .
\end{equation}

In the main experiments, we set $\lambda_{\mathrm{cls}}=0.05$, $\lambda_{\mathrm{reg}}=1.0$, $\lambda_{\mathrm{rank}}=0.05$, and $\lambda_{\mathrm{contact}}=0$. Thus, contact-guided supervision is disabled in the main results.

 The model is trained for 20 epochs using AdamW with weight decay 0.05, batch size 16, and dropout 0.3. The interaction module is optimized with learning rate $1\times10^{-4}$, while the ligand encoder and protein encoder use learning rates $1\times10^{-5}$ and $5\times10^{-6}$, respectively. Both the protein encoder and ligand encoder are fine-tuned, while the bottom 28 layers of ESM are frozen. Layer-wise learning-rate decay is applied with factor 0.97.

The interaction module uses 8 attention heads, sigmoid attention with temperature 0.7, token dropout of 0.10 for both protein and ligand tokens, and bidirectional ligand--protein overlap aggregation. Unless otherwise noted, we use the simple-pooling interaction mode and retain interactions supported from both ligand-to-protein and protein-to-ligand directions. Contact supervision is not used in the main experiments ($\lambda_{\mathrm{contact}}=0$). Models are selected based on validation performance and evaluated once on the held-out test set. All experiments are repeated with 5 random seeds. Hyperparameters are fixed across experiments unless otherwise noted.

We formulate DTI prediction as a binary classification task. Continuous KIBA scores are binarized using a threshold of 12.1, and the model outputs a scalar logit that is converted to an interaction probability by a sigmoid function. The model is trained using a binary cross-entropy loss with logits, with an auxiliary regression loss weighted by $\lambda=0.2$.
\begin{equation}
\mathcal{L}_{\mathrm{DTI}} = - \mathbb{E}_{(x,y)} \left[ y \log \sigma(z) + (1-y)\log(1-\sigma(z)) \right],
\end{equation}
where $z$ is the predicted logit and $\sigma$ denotes the sigmoid function. We train the DTI model for 30 epochs using AdamW with learning rate $3\times10^{-5}$, weight decay 0.05, batch size 16, and dropout 0.4. Both the protein encoder and ligand encoder are fine-tuned, while the bottom 28 layers of ESM are frozen. Layer-wise learning-rate decay is applied with factor 0.97, and the ESM learning-rate multiplier is set to 0.1. We use a linear warmup with ratio 0.05, followed by a plateau-based learning rate scheduler. The interaction module uses 8 attention heads, sigmoid attention with temperature 1.5, token dropout of 0.10 for both protein and ligand tokens, and bidirectional ligand--protein aggregation. Models are selected based on validation NDCG@10 and evaluated once on the held-out test set.
The same training configuration is used for both SMILES and VQ-Atom representations.
We evaluate the model in inference mode without dropout and report test performance using the checkpoint selected on the validation set.
All experiments are repeated with 5 random seeds. Hyperparameters are fixed across experiments unless otherwise noted.

\section{Experiments}

\subsection{Dataset and Evaluation Setup}

We evaluate on the KIBA dataset~\cite{he2017simboost}, following the preprocessing protocol of GraphDTA~\cite{nguyen2021graphdta}. Continuous interaction scores are binarized using a threshold of 12.1. We construct a strict protein-cold split using MMseqs2 clustering~\cite{steinegger2017mmseqs2} at 40\% sequence identity, ensuring that proteins are completely disjoint across train, validation, and test sets. We focus on the protein-cold setting, which is commonly used in prior DTI studies and reflects generalization to unseen targets. The dataset sizes are 35,747 / 4,219 / 4,036 pairs, respectively, with positive rates of 32.5\%, 22.4\%, and 28.9\%.

The number of unique proteins is 113 / 14 / 15, and the number of unique ligands is 1,007 / 841 / 793 across train, validation, and test sets, respectively. While proteins are strictly disjoint, ligands are largely shared across splits. This setup prevents trivial memorization of protein identity while enabling controlled evaluation of ligand representations. The dataset is relatively dense on the protein side, while ligands are sparse in validation and test, creating a challenging generalization setting. Under this structure, each ligand is associated with a limited number of proteins, making ligand–target interactions relatively sparse from the ligand perspective.

We also experimented with stricter double-cold splits, where both proteins and ligands are unseen during testing. However, such splits resulted in extremely small and unstable evaluation sets. In particular, the number of interaction pairs per protein or per ligand often approaches one, making it difficult for models to learn meaningful patterns and leading to high variance in evaluation.

We do not use large-scale datasets such as BindingDB~\cite{liu2007bindingdb} or structure-based datasets such as PDBBind~\cite{wang2004pdbbind} for supervised evaluation. BindingDB is highly sparse, with very few interaction pairs per protein–ligand pair, while PDBBind is relatively small and structurally biased. These characteristics make it difficult to construct reliable protein-cold or double-cold splits and often lead to unstable training and evaluation. Instead, we focus on the KIBA dataset, which provides a denser interaction matrix and a more suitable benchmark for controlled evaluation of representation quality under distribution shift.

Because our primary objective is virtual-screening style ranking rather than binary classification alone, model checkpoints are selected using validation NDCG@10 rather than validation AUROC. Test metrics are reported using the checkpoint that achieves the highest validation NDCG@10.

To isolate the contribution of molecular tokenization from that of the interaction architecture, we also evaluate a simplified interaction module as an ablation study. In this variant, protein and ligand embeddings generated by the Transformer encoders are independently pooled using mean and max operations, and the pooled representations are concatenated before classification. Because this architecture contains no token-level interaction mechanism, performance differences primarily reflect the quality of the underlying molecular representation and its compatibility with Transformer-based encoding. We additionally evaluate a richer interaction module that explicitly models protein--ligand interactions, which serves as the practical configuration used in our final system.

\begin{table*}[t]
\centering
\caption{Comparison of molecular representations on the KIBA protein-cold split.
Results are reported as mean values over five random seeds. BestEpoch denotes the epoch selected by the validation NDCG@10 criterion.}
\label{tab:kiba_main}
\small
\begin{tabular}{lcccccccc}
\toprule
Method &
NDCG@10 &
Best@1\% &
Best@5\% &
Top1 &
EF1 &
EF5 &
AUROC &
BestEpoch \\
\midrule
SMILES
& 0.318 & 0.000 & 0.000 & 0.000 & 0.826 & 0.914 & 0.489 & 1 \\

SMILES+Pretrain
& 0.317 & 0.000 & 0.000 & 0.000 & 1.214 & 1.153 & 0.499 & 1 \\

Continuous
& 0.582 & 0.200 & 0.360 & 0.107 & 3.068 & 2.850 & 0.738 & 29 \\

Continuous+Pretrain
& 0.588 & 0.200 & 0.360 & 0.133 & 3.135 & 2.816 & 0.742 & 31 \\

VQ-Atom
& \textbf{0.627}
& 0.213
& 0.413
& 0.147
& \textbf{3.321}
& \textbf{3.001}
& 0.777
& 10 \\

VQ-Atom+Pretrain
& 0.625
& \textbf{0.280}
& \textbf{0.427}
& \textbf{0.227}
& 3.236
& 2.956
& \textbf{0.788}
& 10 \\
\bottomrule
\end{tabular}
\end{table*}

\subsection{Main Results}

Several observations emerge from Table~\ref{tab:kiba_main}.

First, VQ-Atom with MLM pretraining achieves the best overall performance across nearly all evaluation metrics, including NDCG@10, Best@1\%, Best@5\%, Top1, EF1, and AUROC. This indicates that combining semantic discretization with Transformer pretraining is the most effective representation strategy among those evaluated.

Second, the effect of pretraining differs substantially across representations. Continuous atom representations show only marginal improvements after MLM pretraining, suggesting that pretraining is less effective when the underlying representation remains continuous.

Third, although VQ-Atom achieves the strongest overall performance, the additional benefit obtained from MLM pretraining is relatively modest. This suggests that a substantial fraction of the useful chemical structure may already be encoded by the tokenization itself.

Finally, SMILES performs poorly regardless of whether pretraining is applied. The negligible difference between SMILES and SMILES+Pretrain suggests that pretraining alone is insufficient when the tokenization itself does not provide chemically meaningful units. Together, these results highlight that token quality plays a central role in determining the effectiveness of Transformer-based molecular learning.

\begin{figure}[t]
\centering
\includegraphics[width=0.6\linewidth]{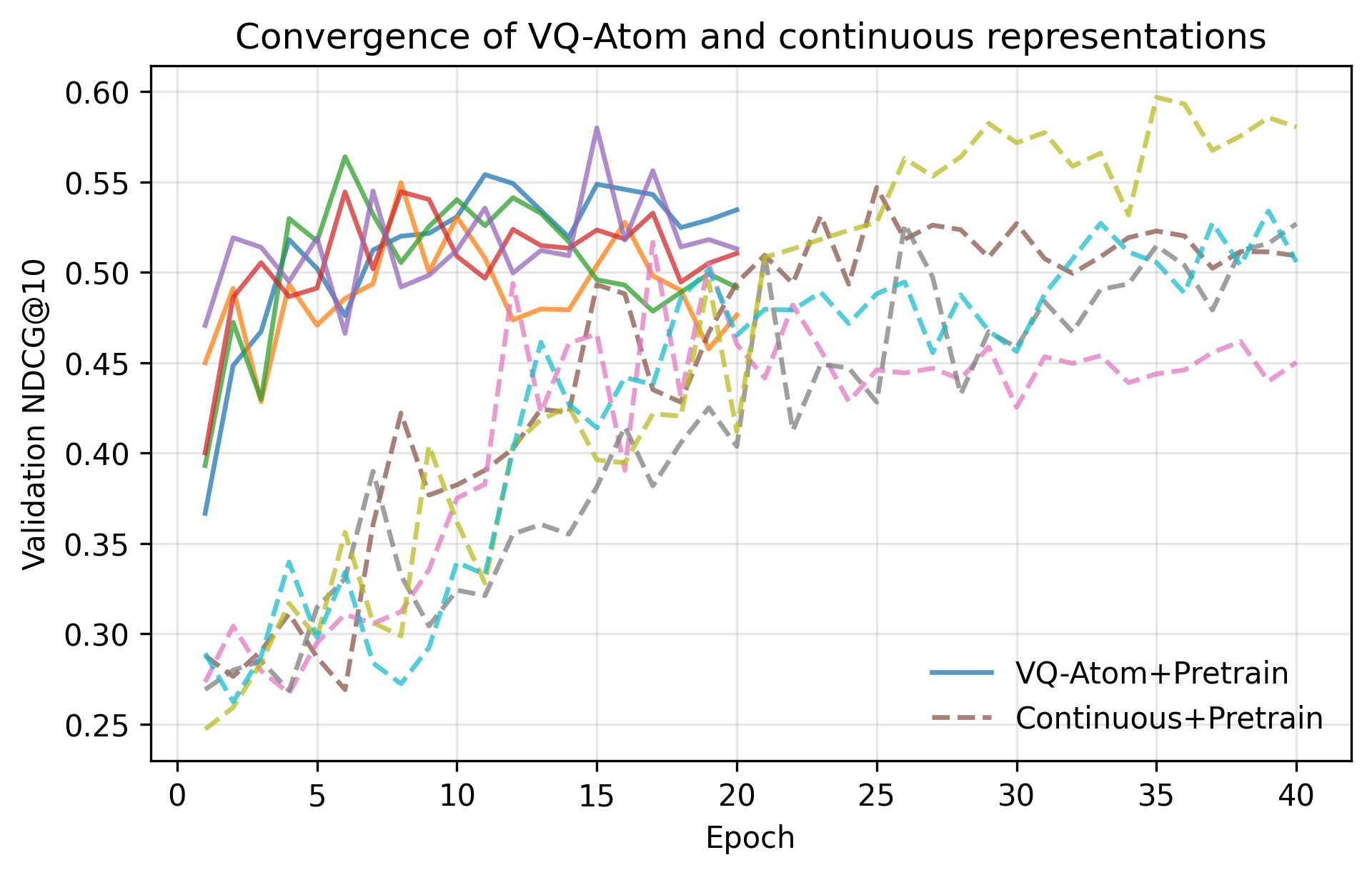}
\caption{
Validation NDCG@10 during downstream DTI training for VQ-Atom and continuous molecular representations across multiple random seeds. Solid lines denote VQ-Atom runs, while dashed lines denote continuous representations. VQ-Atom consistently reaches strong ranking performance within the first 5--10 epochs, whereas continuous representations require substantially more training epochs to achieve comparable performance. This faster convergence suggests that semantic discretization provides a more optimization-friendly representation for Transformer-based molecular learning.
}
\label{fig:convergence}
\end{figure}
Another notable observation is that VQ-Atom converges substantially faster than continuous representations. Across multiple runs, the best VQ-Atom checkpoints were typically obtained within the first 5--10 epochs, whereas continuous representations often required more than 30 epochs to reach comparable performance. This behavior suggests that semantic discretization improves optimization efficiency by exposing recurring chemical environments as reusable symbolic units, making pretrained knowledge easier to exploit during downstream learning.

Figure~\ref{fig:convergence} further illustrates this difference in optimization dynamics. While both representations eventually achieve similar ranking performance, VQ-Atom reaches high NDCG@10 much earlier during training. The reduced number of training epochs required to obtain competitive checkpoints may also translate into lower computational cost in practical applications.

\subsection{Ablation: Tokenization and Interaction Module}

Table~\ref{tab:ablation} compares Continuous and VQ-Atom representations under simplified and full interaction modules. The full cross-attention module improves both representations, confirming that interaction modeling is important for DTI prediction. However, the main observation is that VQ-Atom remains stronger than the Continuous representation under both interaction settings.

Under the simplified interaction module, VQ-Atom+MLM achieves NDCG@10 of 0.525 and AUROC of 0.740, compared with 0.323 and 0.514 for Continuous+MLM. This large gap indicates that VQ-Atom tokens carry useful chemical information even when the interaction module is weakened. With the full interaction module, VQ-Atom+MLM further improves to NDCG@10 of 0.625 and AUROC of 0.788, outperforming Continuous+MLM across all reported metrics.

These results suggest that VQ-Atom does not merely benefit from a stronger downstream architecture. While the full interaction module improves both representations, the Continuous representation depends heavily on the cross-attention design, achieving only 0.323 NDCG@10 and 0.514 AUROC under the simplified interaction setting. In contrast, VQ-Atom retains strong performance even with the simplified interaction module, reaching 0.525 NDCG@10 and 0.740 AUROC.

This distinction is important because the interaction module is not part of the ligand Transformer itself. The large performance drop observed for Continuous representations suggests that much of the predictive signal is recovered by the downstream interaction module rather than by the ligand representation. By contrast, VQ-Atom already provides a strong ligand representation before applying the full interaction design.

These findings indicate that the advantage of VQ-Atom originates primarily from the tokenized molecular representation and the Transformer encoder built upon it, rather than from the downstream interaction module. The full interaction design further improves performance, but VQ-Atom remains substantially stronger than Continuous representations even under a simplified interaction setting. This suggests that semantic discretization produces more effective Transformer-compatible molecular representations than continuous atom-level features.

\begin{table}[t]
\centering
\caption{Ablation of tokenization and cross-attention design on the KIBA protein-cold benchmark.}
\label{tab:ablation}
\small
\begin{tabular}{lcccc}
\hline
Method(All pretrained) & NDCG@10 $\uparrow$ & EF1 $\uparrow$ & EF10 $\uparrow$ & AUROC $\uparrow$ \\
\hline
Continuous + MLM (Simple Cross) & 0.323 & 1.096 & 1.018 & 0.514 \\
Continuous + MLM (Full Cross)   & 0.588 & 3.135 & 2.504 & 0.742 \\
\hline
VQ-Atom + MLM (Simple Cross)    & 0.525 & 2.866 & 2.557 & 0.740 \\
VQ-Atom + MLM (Full Cross)      & \textbf{0.625} & \textbf{3.236} & \textbf{2.735} & \textbf{0.788} \\
\hline
\end{tabular}
\end{table}

\subsection{Why VQ-Atom Tokens Improve Generalization}

The advantage of VQ-Atom can be understood from the role of the token itself. In SMILES, tokens are syntactic symbols derived from a traversal of the molecular graph. In continuous graph representations, each atom is represented by a real-valued vector whose meaning depends on the learned embedding space. In contrast, VQ-Atom assigns each atom to a discrete token corresponding to a local chemical environment. Thus, the token identity itself encodes reusable chemical context.

This has two consequences. First, chemically similar local environments can share the same token across different molecules. The model can therefore accumulate statistical evidence for recurring atom-level motifs, rather than learning each molecular instance independently. Second, discrete tokens provide stable symbolic anchors for Transformer-based modeling. Attention layers can repeatedly encounter the same token IDs across different molecular contexts, allowing the model to learn interaction tendencies associated with chemically meaningful environments.

This explains why VQ-Atom improves ranking-oriented metrics. Protein-cold evaluation requires generalization to unseen proteins, so memorizing protein identity is not possible. A ligand representation that exposes reusable local chemical motifs can help the model rank candidate interactions more consistently. The improvement over Continuous representations suggests that discretization is not simply a compression step; it changes the learning problem by converting continuous atom environments into a molecular vocabulary.

Therefore, the central contribution of VQ-Atom is not only that it uses graph-derived atom features, but that it transforms those features into semantic discrete units. This supports the view that tokenization is a core modeling decision in molecular machine learning, rather than a preprocessing detail.

\subsection{Why Is the Pretraining Gain Marginal?}

An interesting observation is that pretraining provides only modest improvements across all molecular representations. While VQ-Atom consistently outperforms continuous representations and SMILES, the performance gap between pretrained and non-pretrained variants remains relatively small.

One possible explanation is that a substantial portion of the chemically relevant local structure is already captured during the tokenization stage. In VQ-Atom, each token represents a local atomic environment aggregated from the molecular graph. Consequently, short-range chemical relationships, which are often critical for molecular recognition and binding, are partially encoded before Transformer pretraining is performed.

Masked language modeling pretraining is expected to learn both local and long-range dependencies among molecular tokens. However, if the downstream DTI task primarily depends on local chemical environments, much of the information required for prediction may already be available in the token representation itself. Under this hypothesis, pretraining mainly contributes additional higher-order or long-range contextual information, leading to only incremental gains during downstream finetuning.

A similar phenomenon may explain the limited benefit of pretraining for continuous representations. The downstream DTI dataset contains a sufficiently large number of interaction pairs, allowing the model to learn task-relevant molecular patterns directly during supervised training. As a result, the additional information acquired during self-supervised pretraining provides only a modest improvement.

These observations suggest that the quality of the underlying molecular representation may be more important than the presence or absence of pretraining. In particular, VQ-Atom achieves strong performance even without pretraining, indicating that semantically meaningful tokenization itself contributes substantially to downstream learning effectiveness.

\section{Discussion}
Traditional machine learning research focuses primarily on three components:

\begin{enumerate}
\item Model architecture
\item Objective function
\item Optimization algorithm
\end{enumerate}

However, large language models have revealed a fourth component:

\begin{enumerate}
\setcounter{enumi}{3}
\item Token design
\end{enumerate}

A Transformer cannot learn statistical regularities that are not exposed through its tokenization. Consequently, token design directly constrains the information available to downstream learning.

Our results suggest that molecular tokenization is not merely an implementation detail. Under an identical downstream architecture, replacing SMILES with VQ-Atom dramatically changes the effectiveness of pretraining and downstream generalization. This indicates that the design of the token vocabulary itself constitutes an independent learning problem.

Our experiments suggest that effective Transformer learning requires both granularity and discreteness.

Granularity allows the representation to capture fine local structure, whereas discreteness enables repeated statistical accumulation across examples.

SMILES provides discreteness but insufficient granularity. Continuous atom embeddings provide granularity but lack reusable symbolic units. VQ-Atom combines both properties and consequently benefits most from Transformer training. The progression from SMILES to Continuous representations and finally to VQ-Atom (Figure~\ref{fig:granularity}) can be interpreted as moving toward representations that simultaneously maximize granularity and symbolic reusability.

\section{Toward Token Design Science}

A central finding of this work is that tokenization substantially influences the effectiveness of Transformer-based molecular learning. Across identical downstream architectures, changing only the molecular representation led to large differences in ranking performance, convergence speed, and the ability to benefit from pretraining. These results suggest that the choice of tokens is not merely a preprocessing decision but a fundamental component of the learning system itself.

This observation motivates a broader hypothesis. In many domains, the performance of Transformer models may depend not only on model architecture and optimization, but also on how raw inputs are partitioned into symbolic units. The tokenization process determines which patterns become reusable, which relationships are preserved, and ultimately what statistical structure is accessible to the model.

The success of large language models is often attributed to model scale and optimization. However, language models also rely on token systems such as words, subwords, and byte-pair encodings. From the perspective of representation granularity, these token systems may themselves be an important part of why Transformer-based learning is effective.

More generally, we hypothesize that many domains contain latent symbolic units that are not directly observable. The objective of token design is to discover discrete units that simultaneously satisfy two requirements:

\begin{itemize}
\item semantic coherence
\item statistical reusability
\end{itemize}

From this perspective, VQ-Atom represents one instance of a broader class of methods aimed at discovering task-appropriate vocabularies.

Potential future applications include:

\begin{itemize}
\item proteins (VQ-Amino)
\item materials science
\item reaction prediction
\item biological networks
\item natural language retokenization
\end{itemize}

In each case, the objective is not merely compression but the discovery of reusable semantic units that improve downstream learning.
An interesting observation from our experiments is that VQ-Atom receives only a modest benefit from MLM pretraining despite achieving the strongest overall performance. One possible explanation is that VQ-Atom already incorporates substantial local relational information before Transformer training. Each token is constructed from an atom-centered chemical environment that captures neighboring atoms, bond structure, aromaticity, ring information, and other local chemical context. Because many chemically relevant interactions are determined primarily by local environments, a significant portion of the information that would otherwise need to be learned during pretraining may already be encoded in the tokenization itself.

This observation suggests a broader hypothesis regarding retokenization. The extent to which token design can replace or reduce pretraining may depend on how much of the domain's semantics are contained within local relationships. Chemistry is a particularly favorable case because local atomic environments often determine molecular behavior. In domains where meaning is dominated by long-range dependencies, tokenization alone may be insufficient to capture the relevant structure, and large-scale pretraining may remain essential.

Therefore, we do not claim that semantic retokenization universally replaces pretraining. Rather, we hypothesize that token design and pretraining occupy complementary roles. Semantic tokenization may shift part of the learning burden from model parameters into the vocabulary itself, with the magnitude of this effect depending on the locality of the underlying domain. Understanding this trade-off may represent an important direction for future research in token design science.

The success of modern NLP may also be interpreted through the lens of token granularity. Unlike molecules, natural language already possesses human-defined symbolic units such as words and morphemes that approximately correspond to semantic concepts. From this perspective, part of the success of language models may stem from the fact that natural language has already undergone a form of semantic discretization before model training begins.

However, existing NLP tokenizations remain largely handcrafted or frequency-driven. Subword vocabularies such as BPE are optimized primarily for compression and coverage rather than semantic consistency. As a result, phenomena such as polysemy, where a single token represents multiple meanings depending on context, remain common. This observation suggests that there may still be room for learned retokenization approaches that discover alternative vocabularies better aligned with semantic structure. Similar to VQ-Atom in chemistry, vector quantization or related methods may provide a mechanism for automatically refining token granularity in natural language and other domains.

A related caveat is that the effectiveness of token design may depend strongly on the nature of the target task. VQ-Atom is particularly successful in chemistry because many molecular properties are governed by local atomic environments. In such settings, semantic discretization can capture a substantial portion of the relevant structure before downstream learning begins. However, not all domains exhibit this property. For tasks whose behavior depends primarily on long-range relationships rather than local patterns, retokenization may provide limited benefits. In these cases, most of the required information must still be acquired through large-scale pretraining or task-specific optimization. Therefore, we view token design not as a universal replacement for representation learning, but as a complementary mechanism whose effectiveness depends on the locality structure of the underlying domain.

\section{Conclusion}

We introduced VQ-Atom, a semantic discretization framework that converts atom-centered chemical environments into reusable discrete tokens. Through controlled comparisons with SMILES and continuous atom-level representations under an identical downstream architecture, we found that molecular tokenization has a substantial impact on Transformer-based learning and downstream generalization.

Our experiments reveal three key findings. First, SMILES tokenization performs poorly despite providing discrete symbols, suggesting that discreteness alone is insufficient. Second, continuous atom-level representations substantially improve performance, highlighting the importance of fine-grained chemical information. Third, VQ-Atom achieves the strongest overall performance, indicating that combining granularity with semantic discretization yields more effective Transformer-compatible representations.

Interestingly, the gains obtained from VQ-Atom are considerably larger than those obtained from MLM pretraining itself. In our experiments, VQ-Atom without pretraining already outperformed both SMILES and continuous representations, while MLM provided only modest additional improvements. This observation suggests that part of the statistical structure normally learned during pretraining may instead be captured directly through the tokenization process. Because VQ-Atom incorporates local chemical relationships into the token definition, chemically similar environments collapse into shared symbolic units that can be reused across molecules.

More broadly, our results support the view that tokenization is not merely a preprocessing step but a fundamental modeling decision. Traditional machine learning research has focused primarily on model architectures, objective functions, and optimization algorithms. Our findings suggest that representation granularity and token design constitute an additional axis of improvement that can substantially alter what statistical structure is accessible to a model. Semantic tokenization may shift part of the learning burden from model parameters into the vocabulary itself.

We therefore view VQ-Atom as one instance of a broader research direction that we call \emph{Token Design Science}: the study of discovering semantic, reusable symbolic units for machine learning. While chemistry provides a particularly favorable setting because many important relationships are local, the same principles may extend to proteins, materials, biological networks, reaction systems, and potentially natural language itself. 

We believe that understanding the interplay between token design, pretraining, and downstream learning will become an increasingly important problem as machine learning systems continue to scale. More broadly, our results suggest that the discovery of reusable semantic symbols may itself represent a fundamental learning problem, complementary to architecture design, objective design, and optimization.

\bibliography{references}

% \appendix
% \section{Additional Implementation Details}

\end{document}